\documentclass[conference]{IEEEtran}
\IEEEoverridecommandlockouts
\usepackage{cite}
\usepackage{amsmath,amssymb,amsfonts}
\usepackage{algorithmic}
\usepackage{graphicx}
\usepackage{textcomp}
\usepackage{xcolor}
\usepackage{longtable}
\usepackage{array}
\usepackage{booktabs}
\usepackage{adjustbox}
\def\BibTeX{{\rm B\kern-.05em{\sc i\kern-.025em b}\kern-.08em
    T\kern-.1667em\lower.7ex\hbox{E}\kern-.125emX}}
\begin{document}

\title{E\textsuperscript{2}CB2former: Effecitve and Explainable Transformer for CB2 Receptor Ligand Activity Prediction\\
% {\footnotesize \textsuperscript{*}Note: Sub-titles are not captured for https://ieeexplore.ieee.org  and
% should not be used}
% \thanks{Identify applicable funding agency here. If none, delete this.}
}

% \author{\IEEEauthorblockN{Anoymous Authors}}
\author{
\IEEEauthorblockN{
1\textsuperscript{st} Jiacheng Xie\IEEEauthorrefmark{2}
}
\IEEEauthorblockA{\textit{University of Alabama at Birmingham}\\
Birmingham, USA\\
turingrain@gmail.com}
\and
\IEEEauthorblockN{
2\textsuperscript{nd} Yingrui Ji\IEEEauthorrefmark{2}
}
\IEEEauthorblockA{\textit{School of Electronic, Electrical and Communication Engineering} \\
\textit{University of Chinese Academy of Sciences}\\
Beijing, China\\
jiyingrui1996@gmail.com}
\and
\IEEEauthorblockN{
3\textsuperscript{rd} Linghuan Zeng
}
\IEEEauthorblockA{\textit{University of Alabama at Birmingham}\\
Birmingham, USA\\
zengl@uab.edu}
\and
\IEEEauthorblockN{
4\textsuperscript{th} Xi Xiao
}
\IEEEauthorblockA{\textit{University of Alabama at Birmingham}\\
Birmingham, USA\\
xxiao@uab.edu}
\and
\IEEEauthorblockN{
5\textsuperscript{th} Gaofei Chen
}
\IEEEauthorblockA{\textit{University of Alabama at Birmingham}\\
Birmingham, USA\\
gchen2@uab.edu}
\and
\IEEEauthorblockN{
6\textsuperscript{th} Lijing Zhu
}
\IEEEauthorblockA{\textit{Bowling Green State University}\\
Bowling Green, USA\\
lijingz@bgsu.edu}
\and
\IEEEauthorblockN{
7\textsuperscript{th} Joyanta Jyoti Mondal
}
\IEEEauthorblockA{\textit{University of Delaware}\\
Newark, USA\\
joyanta@udel.edu}
\and
\IEEEauthorblockN{
8\textsuperscript{th} Jiansheng Chen\IEEEauthorrefmark{1}}
\IEEEauthorblockA{\textit{Aerospace Information Research Institute} \\
\textit{Chinese Academy of Sciences}\\
Beijing, China \\
chenjs@aircas.ac.cn}
\thanks{\IEEEauthorrefmark{2} Equal contribution.}
\thanks{\IEEEauthorrefmark{1} Corresponding author.}
}

\maketitle

\begin{abstract}
Accurate prediction of CB2 receptor ligand activity is pivotal for advancing drug discovery targeting this receptor, which is implicated in inflammation, pain management, and neurodegenerative conditions. Although conventional machine learning and deep learning techniques have shown promise, their limited interpretability remains a significant barrier to rational drug design. In this work, we introduce CB2former, a framework that combines a Graph Convolutional Network (GCN) with a Transformer architecture to predict CB2 receptor ligand activity. By leveraging the Transformer’s self‐attention mechanism alongside the GCN’s structural learning capability, CB2former not only enhances predictive performance but also offers insights into the molecular features underlying receptor activity. We benchmark CB2former against diverse baseline models—including Random Forest, Support Vector Machine, K‐Nearest Neighbors, Gradient Boosting, Extreme Gradient Boosting, Multilayer Perceptron, Convolutional Neural Network, and Recurrent Neural Network—and demonstrate its superior performance with an R\textsuperscript{2} of 0.685, an RMSE of 0.675, and an AUC of 0.940. Moreover, attention‐weight analysis reveals key molecular substructures influencing CB2 receptor activity, underscoring the model’s potential as an interpretable AI tool for drug discovery. This ability to pinpoint critical molecular motifs can streamline virtual screening, guide lead optimization, and expedite therapeutic development. Overall, our results showcase the transformative potential of advanced AI approaches—exemplified by CB2former—in delivering both accurate predictions and actionable molecular insights, thus fostering interdisciplinary collaboration and innovation in drug discovery.
\end{abstract}


\section{Introduction}
Cannabis is one of the most widely used drugs globally, producing a diverse array of pharmacological effects in humans \cite{connor2021cannabis, xiao2025tdrdtopdownbenchmarkrealtime, Atakan2012CannabisAC}. Numerous studies have demonstrated that cannabis and cannabinoids offer significant therapeutic benefits, including stimulating appetite and alleviating nausea and vomiting \cite{robson2014therapeutic, gasperi2023recent, 10825244, DAE127603, hill2015medical}. The discovery of cannabinoid receptors (CBRs) provided insight into the mechanism of action of cannabinoid drugs. CBRs are classified into two types: cannabinoid receptor type 1 (CB1) and cannabinoid receptor type 2 (CB2) \cite{gasperi2023recent, catani2016assay}, both belonging to the G protein-coupled receptor (GPCR) subfamily \cite{citation-0, KATRITCH2012, weis2018themolecular}. Figure 1 illustrates the structure of CB1 and CB2 receptors, which include three extracellular loops (ECL1, ECL2, and ECL3), three intracellular loops (ICL1, ICL2, and ICL3), and seven transmembrane domains (7TM). These domains link the glycosylated extracellular amino-terminal (N-term) and intracellular carboxyl-terminal (C-term) domains, a characteristic shared by all GPCRs.

\begin{figure}[htbp]
    \centering     
    \includegraphics[width=0.5\textwidth]{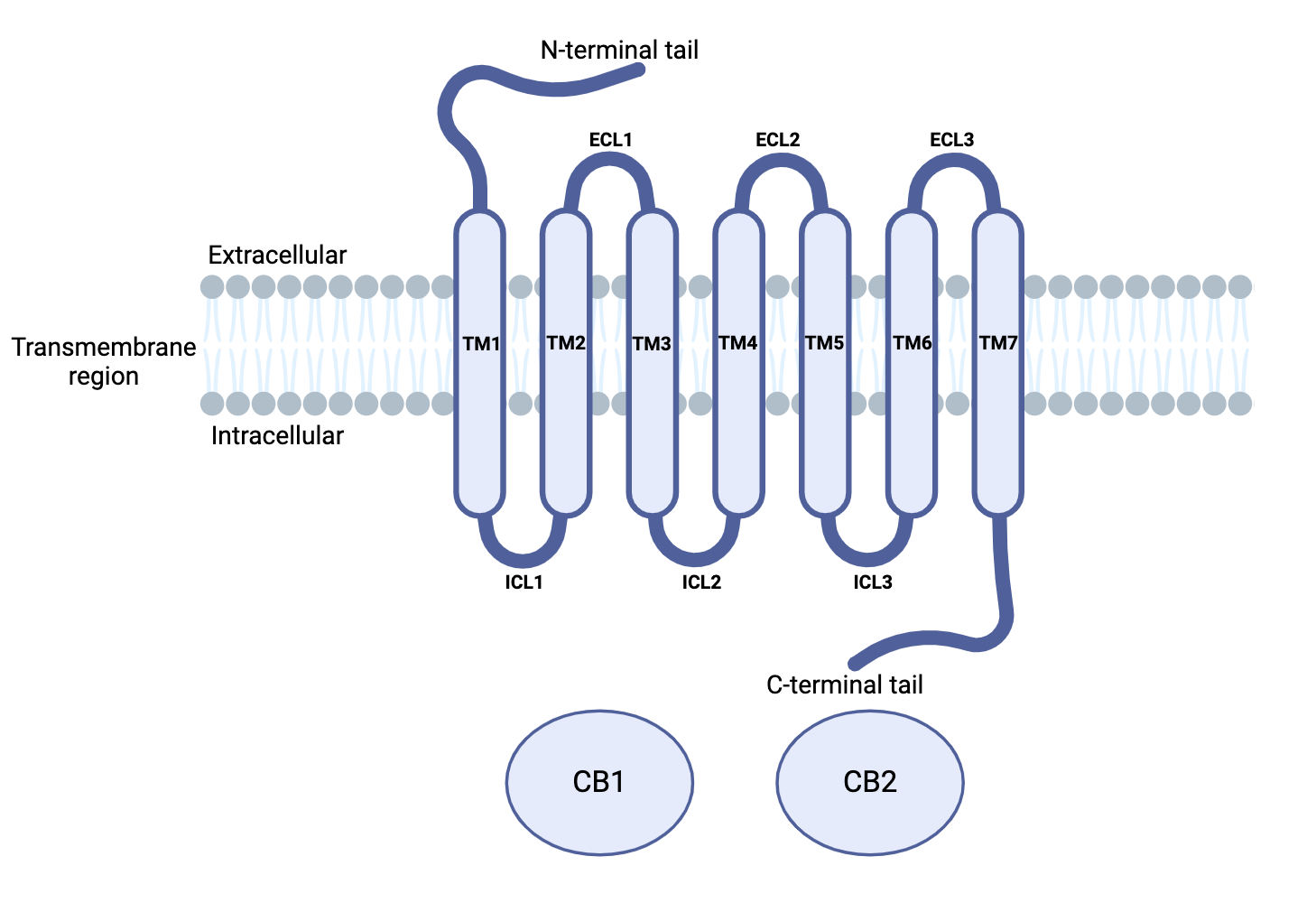}
    % \flushleft
    \caption{Schematic model of CB1 and CB2, showing the terminal tails, TM helices and intracellular loops}
    \label{fig:GPCRs}
\end{figure}

CB1 receptors are predominantly located in the brain, particularly in the cerebral cortex, hippocampus, basal ganglia, and cerebellum \cite{busquets2018cb1}. Activation of CB1 receptors is associated with psychotropic effects \cite{koethe2007expression, long2012distinct}. However, their use was discontinued due to significant side effects on the central nervous system, including anxiety, depression, and suicidal thoughts \cite{ortiz2022medicinal}. These adverse effects overshadowed the potential therapeutic benefits of CB1 activation, leading to its removal from the market and the cessation of further clinical development \cite{finn2021cannabinoids}.

In contrast, CB2 receptors are primarily found in peripheral tissues, particularly in immune-related areas such as the spleen, tonsils, thymus, mast cells, and blood cells \cite{turcotte2016cb}. As a result, CB2 receptors represent a promising therapeutic target for a variety of diseases without inducing psychotropic side effects \cite{aghazadeh2016medicinal}, including conditions such as inflammatory and neuropathic pain \cite{guindon2008cannabinoid,xiao2024hgtdpdtahybridgraphtransformerdynamic, toxins13020117, ijms25116268, ijms21030747}. Consequently, there has been a growing interest in identifying new CB2 ligands \cite{wu2022rational}.

A more affordable and efficient method to accelerate the design and identification of novel CB2 ligands is urgently needed, as traditional experimental screening is both costly and time-consuming \cite{szwabowski2023application, carracedo2021review}. Figure 2 shows the structure of representative CB2 ligands. In the following sections, we will employ algorithms to demonstrate the crucial role these ligands play in CB2 receptor activity.

\begin{figure}[htbp] 
    \centering     
    \includegraphics[width=0.5\textwidth]{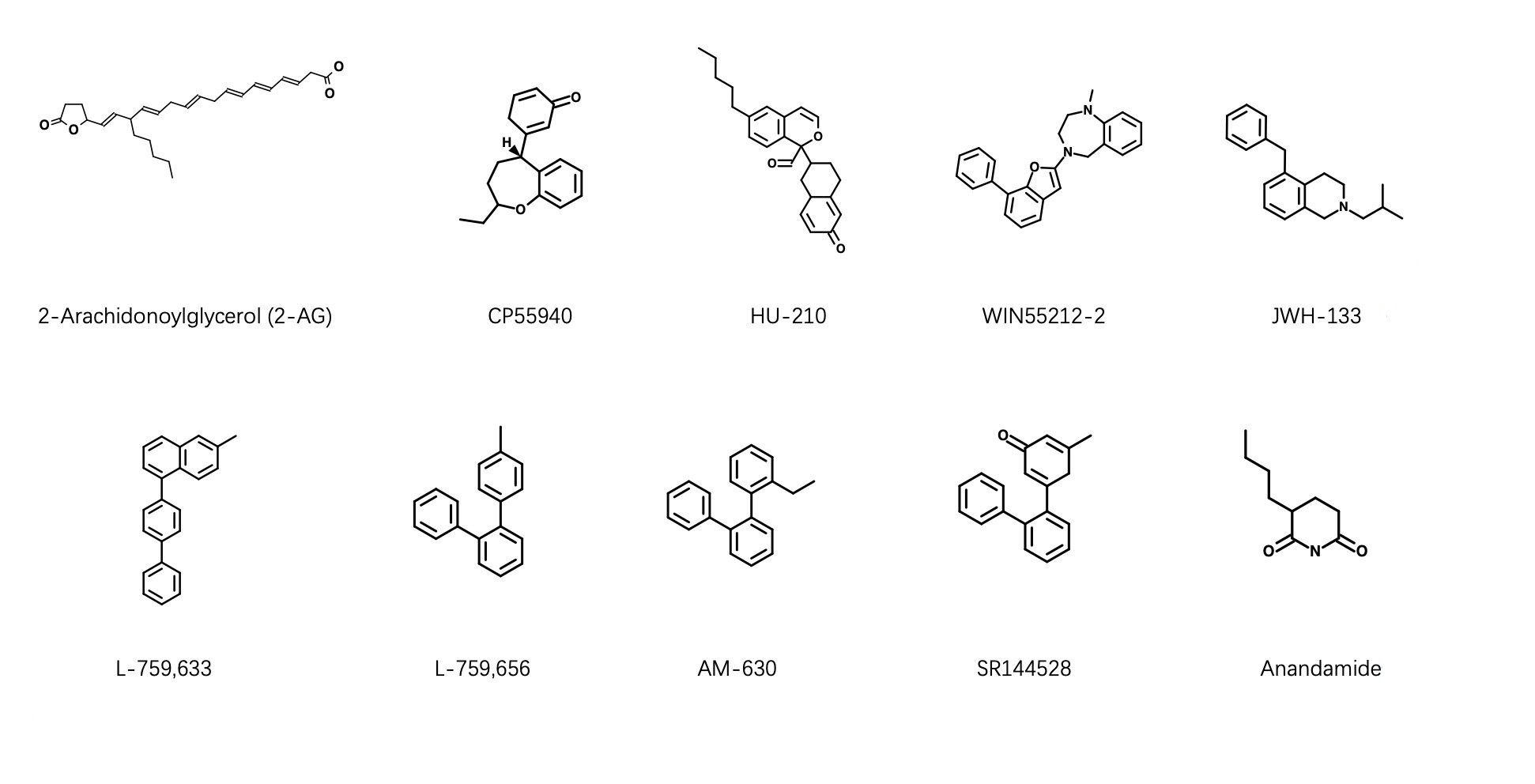}
    % \flushleft
    \caption{Structure of representative CB2 ligands}
    \label{fig:CB2}
\end{figure}

As algorithms continue to evolve and improve, an increasing number of methods are being developed to predict molecular activity. This growing array of computational tools includes a range of approaches such as machine learning algorithms, molecular modeling techniques, and quantitative structure-activity relationship (QSAR) models \cite{ching2018oppor, vama2019application}. These advancements allow researchers to gain deeper insights into the interactions between molecules and their targets, thereby enhancing the prediction of biological activity and guiding drug discovery efforts \cite{hill2015medical, RUSSO2017198}.

For example, in 2012, Myint KZ et al. \cite{myint2012molecular} introduced FANN-QSAR, a novel method for predicting the biological activity of diverse chemical ligands. Their study showed that FANN-QSAR effectively identified CB2 lead compounds with strong binding affinity and compounds resembling known cannabinoids. Furthermore, the model offered insights into variations in R-groups and scaffolds among known ligands. This underscores the importance of meticulous molecular descriptor selection in virtual screening for identifying bioactive molecules. In 2013, Chao Ma et al. \cite{ma2013licabeds} expanded LiCABEDS for cannabinoid ligand selectivity modeling using molecular fingerprints, comparing its performance with SVM. Analysis of LiCABEDS models highlighted structural differences between CB1 and CB2 ligands, aiding in novel compound design. Importantly, they found that LiCABEDS successfully identified newly synthesized CB2 selective compounds, showcasing its potential for drug discovery. In 2017, Cano G et al. \cite{cano2017automatic} used the power of random forests to automatically select features (molecular descriptors), thereby avoiding manual selection of descriptors and improving activity prediction by improving goodness of fit. In 2018, Floresta G et al. \cite{floresta2018discovery} proposed two new 3D-QSAR models that comprehend a large number of chemically different CB1 and CB2 ligands and well account for the individual ligand affinities. These features will facilitate the recognition of new potent and selective molecules for CB1 and CB2 receptors. SVM-based techniques are considered a powerful approach in early drug discovery. In 2020, Stokes JM et al. \cite{stokes2020deep} proposed a deep neural network (DNN) that can predict molecules with antimicrobial activity by using a directed information transfer neural network (D-MPNN) architecture. They successfully identified halicin as a new broad-spectrum antimicrobial compound, demonstrating the effectiveness of deep learning approaches in drug discovery. In 2021, Mukuo Wang et al. \cite{wang2021identification} utilized a "deep learning–pharmacophore–molecular docking" virtual screening strategy to identify CB2 antagonists from the ChemDiv database. Testing 15 hits, seven showed binding affinities against CB2, with Compound 8 exhibiting the strongest activity. The 4H-pyrido[1,2-a] pyrimidin-4-one scaffold in Compound 8 holds promise for CB2 drug development. The study suggests further application of their model for identifying novel CB2 antagonists and designing potent ligands for diverse therapeutic targets. In 2022, Yuan et al. \cite{yuan2022silico} utilized MCCS to analyze CB2 allosteric modulation. Docking known CB2 allosteric modulators (AMs) Ec2la, trans-$\beta$-caryophyllene, and cannabidiol (CBD), they identified potential binding sites. Molecular dynamics simulations suggested site H as most promising. Future plans include bio-assay validations. In 2022, Cerruela-Garcia et al. \cite{cerruela2022graph} proposed a feature selection method based on undirected graph construction and successfully applied it to molecular activity prediction research. In 2022, Irwin R et al.\cite{irwin2022chemformer} proposed the Chemformer model, which makes use of the SMILES language for application to diverse computational [9] chemistry tasks. the Chemformer model can be applied to a wide variety of downstream tasks, including both sequence-to-sequence and discriminative tasks, fairly easily. In 2023, Zhou et al. \cite{zhou2023reliable} demonstrated the efficacy of machine learning in predicting cannabinoid receptor 2 (CB2) ligands. They compared descriptor-based and graph-based models, finding XGBoost to excel in both regression and classification tasks, especially when utilizing specific combinations of molecular fingerprints. Outperforming existing QSAR models, their approach enhances CB2 ligand prediction accuracy, offering insights for novel therapy design. This study advances computational drug discovery, with potential applications across diverse drug targets. The researchers employed varied methodologies and successfully developed effective models. These findings demonstrate the versatility of their approaches and highlight the achievement of acceptable predictive models.

%\textcolor{blue}{In this study, we comprehensively compared various machine learning and deep learning models, including Random Forest, Support Vector Machine, K-Nearest Neighbors, Gradient Boosting Machine, Extreme Gradient Boosting, Multilayer Perceptron, Convolutional Neural Network, Recurrent Neural Network, and the Transformer model, to predict CB2 receptor ligand activity. We aim to demonstrate the superiority and interpretability of the CB2former in this context, contributing to the advancement of AI-driven drug discovery.}

% 增加动态prompt相关内容
In particular, we introduce a \emph{Prompt-based} CB2former approach, wherein CB2-related structural and functional information is injected into the model as unlearnable prompt tokens. This design encodes key receptor knowledge directly into the model’s input sequence, potentially accelerating model convergence and providing a more interpretable focus on critical molecular substructures.

To rigorously evaluate CB2former, we compare it against a comprehensive set of baseline models—including Random Forest, Support Vector Machine, K-Nearest Neighbors, Gradient Boosting Machine, Extreme Gradient Boosting, Multilayer Perceptron, Convolutional Neural Network, and Recurrent Neural Network. Our results demonstrate that CB2former outperforms these established methods, achieving an \textit{R}\textsuperscript{2} of 0.685, an RMSE of 0.675, and an AUC of 0.940 on benchmark datasets. Further, attention‐weight analysis reveals key molecular substructures driving CB2 receptor activity, underscoring the framework’s potential as an interpretable AI tool for drug discovery. By pinpointing critical motifs, researchers can streamline virtual screening, guide lead optimization, and expedite therapeutic development.

The primary objective of this study is thus to develop an advanced, interpretable predictive model for CB2 receptor ligand activity that leverages the synergistic benefits of the Transformer’s self-attention and the GCN’s graph-based structural learning. Our overarching aim is to address the shortcomings of traditional machine learning and deep learning approaches—namely, limited interpretability—by providing a model that delivers both high predictive accuracy and actionable molecular insights.

The key contributions of this study are as follows:

\textbf{Application and Innovation of an CB2former}: We present one of the first frameworks integrating a GCN with a Transformer to predict CB2 receptor ligand activity. By uniting the Transformer’s attention mechanism with graph-based structural learning, CB2former captures complex molecular relationships in SMILES strings and molecular graphs, offering robust predictive power. Notably, the inclusion of unlearnable prompt tokens encoding CB2-related structural knowledge enhances both accuracy and interoperability. 

\textbf{Enhanced Predictive Accuracy and Comprehensive Model Comparison}: A head-to-head comparison with various baselines—ranging from classic machine learning models (Random Forest, Support Vector Machine, K-Nearest Neighbors, Gradient Boosting Machine, Extreme Gradient Boosting) to modern deep learning architectures (Multilayer Perceptron, Convolutional Neural Network, Recurrent Neural Network)—demonstrates the superior performance of CB2former. These results offer a thorough benchmark for the cheminformatics community. 

\textbf{Advancement of AI4Science and Potential for Drug Discovery}: By showcasing how an advanced AI approach can yield both high accuracy and interpretability, this study exemplifies the transformative role of AI in scientific research. Our findings have immediate applications in drug discovery, particularly in virtual screening and lead optimization for CB2-related therapeutics. This work thus highlights the broader impact of integrated AI frameworks in expediting the development of novel treatments for diseases linked to the CB2 receptor.

\section{Materials and Methods}\label{sec2}

\subsection{Data Collection}\label{subsec2.1}

\subsubsection{Data Sources}\label{subsec2.1.1}

The data used in this study were sourced from two comprehensive and publicly accessible chemical databases, ChEMBL and BindingDB, both of which provide detailed information on chemical compounds, their molecular structures, and associated bioactivities. For the purposes of this research, we specifically selected compounds with documented interactions involving the Cannabinoid Receptor 2 (CB2).

\textbf{ChEMBL}: A curated repository of bioactive molecules with drug-like properties, containing extensive bioactivity data derived from the scientific literature. This database is widely recognized for its breadth and accuracy of compound–target associations.

\textbf{BindingDB}: A web-accessible database focused on measured binding affinities, particularly the interactions of proteins deemed drug targets with small, drug-like molecules. 

From these two databases, a total of 1000 compounds with verified CB2 activity were compiled. The collected bioactivity information, including Ki, IC50, and EC50 values, formed the foundation of the dataset used for model training and evaluation.

\subsubsection{Data Preprocessing}\label{subsec2.1.2}

Prior to model development, the raw data underwent several preprocessing steps—data cleaning, normalization, and transformation—to ensure consistency and quality.

\textbf{Data Cleaning}: Duplicate entries and compounds with missing or incomplete bioactivity values were excluded. Additionally, any compounds with ambiguous or inconsistent activity measurements were removed to maintain a high-quality dataset.

\textbf{Normalization}: All bioactivity measures (Ki, IC50, EC50) were converted to a uniform \textit{pIC50} scale, defined as -log10(IC50). This conversion standardizes the range of reported values, facilitating more direct comparisons among different bioactivity measurements.

\textbf{Transformation}: Molecular structures were represented using canonical SMILES (Simplified Molecular Input Line Entry System) notation to ensure each compound’s structure was captured uniquely and consistently. This canonicalization step is critical for maintaining structural uniformity across the dataset and for enabling accurate downstream analyses.

\subsection{Molecular Representation}\label{subsec2.2}

Effective application of machine learning models for CB2 receptor ligand activity prediction requires the transformation of chemical structures into a numerical form. In this study, we employed two complementary molecular representations—SMILES strings and molecular fingerprints—to capture both sequential and substructural information relevant to ligand–receptor interactions.

\subsubsection{SMILES Strings}\label{subsec2.2.1}

SMILES (Simplified Molecular Input Line Entry System) is a compact, text-based format for describing chemical structures. Its human-readable nature and widespread compatibility make SMILES a popular choice for computational chemistry tasks.

In this work, SMILES strings were generated for each compound and used as the primary input to the CB2former. By encoding atomic connectivity and bond information in a linear sequence, SMILES strings enable sequence-based deep learning models—such as the Transformer architecture—to learn meaningful representations of molecular structure and reactivity.

\subsubsection{Molecular Fingerprints}\label{subsec2.2.2}

Molecular fingerprints encode the presence or absence of specific substructures within a molecule into a fixed-length bit vector, facilitating tasks like similarity searching and machine learning classification or regression. Here, we utilized Extended Connectivity Fingerprints (ECFP), also referred to as Morgan fingerprints, owing to their proven effectiveness in capturing local substructural features.

The fingerprints were calculated as follows:
\begin{enumerate}
    \item  Each molecule was converted from its SMILES string into an RDKit molecular graph. 
    \item ECFP fingerprints (radius 2, 2048-bit length) were then generated from this graph, resulting in a binary vector representation that highlights critical substructures.
\end{enumerate}

These fingerprints were primarily used by models that process numerical input, enabling them to associate specific substructural patterns with variations in CB2 receptor activity.

\subsubsection{Chemical Structures of Selected Compounds}\label{subsec2.2.3}

\begin{figure}[!htbp]
    \centering
    \includegraphics[width=0.5\textwidth]{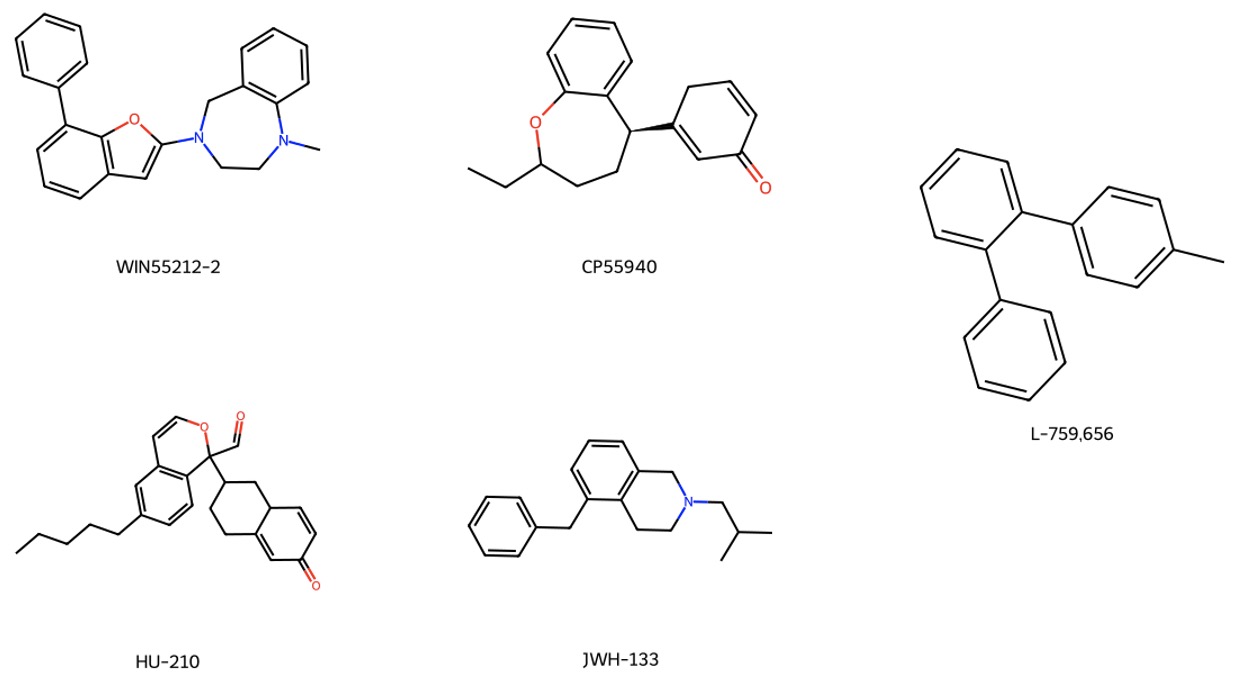}
    \caption{Chemical structures of selected compounds.}
    \label{fig:compound_structures}
\end{figure}

For illustrative purposes, Figure \ref{fig:compound_structures} presents the chemical structures of selected representative compounds from the dataset. Each structure was visualized through RDKit, utilizing the corresponding SMILES string for canonical representation. 
% These examples help underscore the diversity of chemical scaffolds included in this study and highlight the importance of robust molecular representations for accurate activity prediction.

\subsection{Model Development}\label{subsec2.3}

\subsubsection{Baseline Models}\label{subsec2.3.1}

To establish a performance benchmark for predicting CB2 receptor ligand activity, we first implemented several widely recognized machine learning (ML) and deep learning (DL) algorithms. These baseline models serve as a comparative foundation against which the proposed CB2former can be evaluated.

We explored several classical ML algorithms known for their effectiveness in various predictive tasks. These models were evaluated for their performance in predicting the activity of CB2 receptor ligands.

\textit{Random Forest (RF)}: An ensemble learning method that constructs multiple decision trees during training. Predictions are made by aggregating the outputs from all constituent trees, either via majority voting (classification) or averaging (regression). Random Forests are known for their robustness, especially in high-dimensional feature spaces. 

\textit{Support Vector Machine (SVM)}: A supervised learning algorithm effective for both classification and regression tasks. By mapping input data into high-dimensional spaces, SVMs can capture complex relationships, making them well suited for scenarios where the feature space is large relative to the number of samples. 

\textit{K-Nearest Neighbors (KNN)}: A simple, non-parametric method that classifies or regresses based on the \( k \) most similar (nearest) training examples in the feature space. Despite its simplicity, KNN can perform competitively on certain datasets with appropriate distance metrics and parameter tuning. 

\textit{Gradient Boosting Machine (GBM)}: An iterative ensemble technique where each new weak learner—often a decision tree—attempts to correct the errors of the previous ensemble. GBM has the flexibility to handle a variety of data distributions and outliers.

\textit{Extreme Gradient Boosting (XGBoost)}: A highly optimized library built upon the gradient boosting framework. XGBoost emphasizes parallelization, regularization, and efficient hardware utilization, often resulting in superior training speeds and predictive performance.

For deep learning baselines, we employed the following models:

\textit{Multilayer Perceptron (MLP)}: A fully connected feedforward neural network composed of multiple layers of perceptrons. MLPs excel at capturing non-linear relationships, making them suitable for various types of structured numerical inputs such as fingerprints or descriptor vectors.

\textit{Convolutional Neural Networks (CNNs)}: Although CNNs are traditionally applied to 2D image data, they can also process generated molecular images or 1D feature maps (e.g., transformed SMILES strings). Their localized receptive fields enable CNNs to learn relevant substructures and spatial patterns in the data.

\textit{Recurrent Neural Networks (RNNs)}: Designed for sequence-based data, RNNs such as LSTM (Long Short-Term Memory) networks are adept at capturing dependencies within sequential inputs like SMILES strings. This ability to handle long-term context makes them particularly suitable for representing extended molecular sequences.

All baseline models were trained on the preprocessed dataset and tuned via standard hyperparameter optimization strategies. Their predictive performances were evaluated using \emph{R}-squared (R\textsuperscript{2}), Root Mean Squared Error (RMSE), and Area Under the Curve (AUC) to provide a comprehensive assessment of both regression accuracy and classification-like discrimination.

\subsubsection{CB2former}

%\textcolor{blue}{The CB2former, originally designed for natural language processing tasks, has shown great promise in various domains due to its ability to handle sequential data and capture long-range dependencies. In this study, we adapted the Transformer model for molecular property prediction by treating SMILES strings as sequences of characters and integrating a GCN to enhance its capabilities.
%这里放模型结构图
%The CB2former consists of an encoder built using stacked layers of self-attention mechanisms and point-wise, fully connected layers, integrated with a GCN layer for structural learning.}

\begin{figure*}[!htbp]
    \centering
    \includegraphics[width=\textwidth]{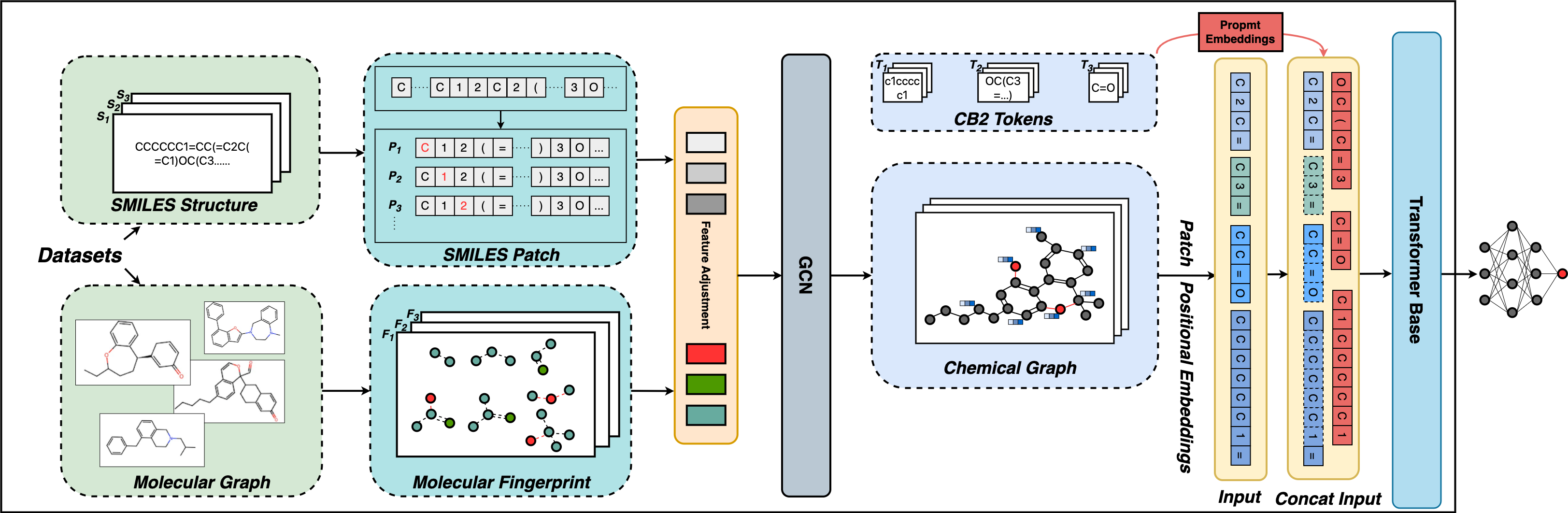}
    \caption{The architecture of The CB2former.}
    \label{fig:architecture}
\end{figure*}

Originally introduced for natural language processing tasks, the Transformer architecture has demonstrated remarkable flexibility in handling sequential data and capturing long-range dependencies. Figure \ref{fig:architecture} illustrates our adaptation of this architecture—referred to as CB2former—for molecular property prediction. In addition to treating SMILES strings as sequential input, we introduce a dynamic Prompt mechanism to incorporate CB2 receptor–specific knowledge directly into the model.

To guide the model toward receptor-specific features, we incorporate a set of unlearnable Prompt tokens—each embedding prior knowledge about CB2‐related functional groups, known binding motifs, or critical substructures—directly into the input. These tokens are concatenated with the SMILES embeddings before being fed into the Transformer, thereby highlighting receptor-relevant information during the attention process. In parallel, a Graph Convolutional Network (GCN) layer is integrated to capture topological insights from molecular graphs, further enhancing the model’s representational capacity.

The CB2former is composed of an encoder that combines stacked layers of domain knowledge (via prompt tokens) and self-attention, interleaved with point-wise, fully connected (feed-forward) layers. The key components are:

\begin{itemize} 
    \item Graph Convolutional Network (GCN) Layer: Processes the molecular graphs to learn representations of each atom within the context of its neighbors. This structural understanding complements the sequence-based encoding from SMILES. 
    \item Embedding Layer: Converts the SMILES tokens into dense vector representations. The unlearnable Prompt tokens, carrying CB2-specific knowledge, are also embedded into similar dimensions and concatenated with these SMILES embeddings. \item Positional Encoding: Injects information about the token positions into the embeddings, ensuring that the model preserves the order of SMILES tokens. 
    \item Self-Attention Layers: Enable the model to focus on different parts of the input sequence dynamically. Multiple attention heads capture diverse representation subspaces, guided by the prior knowledge encoded in the Prompt tokens. 
    \item Feed-Forward Layers: Apply non-linear transformations to the outputs of the self-attention layers, further refining the learned representations.
 \end{itemize}
By combining sequential (SMILES-based) and structural (GCN-based) insights, the CB2former aims to provide a more holistic view of each molecule. Notably, the injection of receptor-specific prior knowledge can accelerate model convergence and improve interpretability, as the self-attention mechanism is guided toward critical features of the CB2 receptor.

A variety of hyperparameters were optimized through an extensive search using cross-validation. The key hyperparameters included: Number of encoder layers, Dimensionality of embeddings and hidden states, Number of attention heads in each self-attention layer, Dimensionality of the feed-forward subnetwork and Dropout rate applied for regularization. 

The best-performing configuration was chosen based on validation performance, striking a balance between model capacity and generalization.

\subsubsection{Self-Attention Mechanism}\label{subsec2.3.3}

The self-attention mechanism is a core component of the Transformer model, allowing it to dynamically focus on different parts of the input sequence. This mechanism is particularly useful for capturing the relationships between atoms in a molecule, which are crucial for accurate molecular property prediction, such as predicting the activity of CB2 receptor ligands.

\paragraph{Attention Mechanism Details}

In the self-attention mechanism, each token in the input sequence is represented by three vectors: Query (Q), Key (K), and Value (V). The attention score for each pair of tokens is calculated using the dot product of the Query and Key vectors, scaled by the square root of the dimensionality of the Key vectors, and passed through a softmax function to obtain the attention weights. This process is particularly beneficial for understanding the importance of different molecular features in predicting CB2 receptor activity.

\[
\text{Attention}(Q, K, V) = \text{softmax}\left(\frac{QK^T}{\sqrt{d_k}}\right)V
\]

Where \( d_k \) is the dimensionality of the Key vectors. The attention weights are then used to compute a weighted sum of the Value vectors, effectively allowing the model to aggregate information from different parts of the sequence. This aggregation helps identify which atoms or substructures within a molecule are most relevant for binding to the CB2 receptor, providing insights into the molecular interactions that drive biological activity.

\paragraph{Multi-Head Attention}

To enhance the model's ability to capture different types of relationships, the self-attention mechanism is extended to multiple heads, each operating in a different subspace of the input representations. The outputs of all attention heads are concatenated and linearly transformed to produce the final output. This multi-head approach allows the model to attend to various aspects of the molecular structure simultaneously, increasing the robustness and accuracy of predictions for CB2 receptor ligands.

\[
\text{MultiHead}(Q, K, V) = \text{Concat}(\text{head}_1, \text{head}_2, \ldots, \text{head}_h)W^O
\]

Where each attention head \( \text{head}_i \) is computed as:

\[
\text{head}_i = \text{Attention}(QW_i^Q, KW_i^K, VW_i^V)
\]

Here, \( W_i^Q \), \( W_i^K \), and \( W_i^V \) are the learned projection matrices for the Query, Key, and Value vectors of the \( i \)-th head, and \( W^O \) is the projection matrix for the concatenated outputs.

\paragraph{Relevance to CB2 Receptor Ligand Prediction}

The self-attention mechanism's ability to highlight important molecular features makes it particularly relevant for predicting CB2 receptor ligand activity. By analyzing the attention weights, we can identify which parts of the molecule contribute most significantly to its binding affinity and activity. This interpretability not only enhances the predictive performance but also provides valuable insights for drug discovery, enabling the design of new compounds with optimized properties. The ability to understand and visualize the key interactions within the molecular structure is a significant innovation in this study, leveraging the power of explainable AI to advance scientific research in cheminformatics and drug discovery.

\subsection{Model Evaluation}\label{subsec2.5}

\subsubsection{Evaluation Metrics for Regression}\label{subsec2.5.1}

To evaluate the performance of regression models predicting the activity of CB2 receptor ligands, the following metrics were used:

\textbf{R-squared (\(R^2\))}: This metric measures the proportion of variance in the dependent variable that is predictable from the independent variables. It is calculated as:
\[
R^2 = 1 - \frac{SS_{res}}{SS_{tot}}
\]
where \(SS_{res}\) is the sum of squares of residuals, and \(SS_{tot}\) is the total sum of squares.

\textbf{Root Mean Squared Error (RMSE)}: This metric measures the average magnitude of the errors between predicted and observed values. It is calculated as:
\[
RMSE = \sqrt{\frac{1}{n} \sum_{i=1}^n (y_i - \hat{y}_i)^2}
\]
where \(n\) is the number of observations, \(y_i\) is the actual value, and \(\hat{y}_i\) is the predicted value.

\subsubsection{Evaluation Metrics for Classification}\label{subsec2.5.2}

For classification tasks, the models' performance was evaluated using the following metrics:

\textbf{Area Under the Curve (AUC)}: This metric measures the ability of the model to distinguish between classes. It is calculated as the area under the Receiver Operating Characteristic (ROC) curve, which plots the true positive rate against the false positive rate at various threshold settings.

\textbf{Accuracy}: This metric measures the proportion of correctly classified instances among the total instances. It is calculated as:
\[
Accuracy = \frac{TP + TN}{TP + TN + FP + FN}
\]
where \(TP\) is true positives, \(TN\) is true negatives, \(FP\) is false positives, and \(FN\) is false negatives.

\subsection{Feature Importance and Interpretation}\label{subsec2.6}

A clear understanding of which features most significantly influence model predictions is critical for interpretability—particularly in scientific research, where insights into molecular mechanisms can guide subsequent experimental and therapeutic strategies.

\subsubsection{SHAP Analysis}\label{subsec2.6.1}

To quantify feature contributions in a model‐agnostic fashion, we employ SHapley Additive exPlanations (SHAP). Grounded in cooperative game theory, SHAP assigns each feature a value that indicates its contribution to the deviation of a given prediction from the model’s mean prediction. SHAP values offer a standardized approach to feature importance across diverse model architectures, including traditional ML models (e.g., RF, SVM) and deep learning frameworks (e.g., MLP, CNN). By examining SHAP values for individual predictions, one can identify the specific features that drive a particular result. Aggregating these values across the dataset provides a global perspective on which features consistently exert the greatest influence. SHAP value plots and summary diagrams highlight which features shift model outputs the most. This level of transparency can reveal hidden patterns and elucidate molecular descriptors that are particularly relevant for CB2 receptor activity.

\subsubsection{Attention Weight Analysis}\label{subsec2.6.2}

In addition to SHAP, the \textbf{self‐attention mechanism} within the CB2former provides an inherently interpretable pathway for understanding how different regions of a molecule contribute to its predicted activity. During a forward pass, the model calculates attention distributions that indicate the relative importance of each token—representing atoms or substructures—in the SMILES sequence. By visualizing these attention weights as heatmaps, one can pinpoint which parts of the molecule the model deems most pertinent for CB2 receptor binding. Since attention weights are learned directly from training data, they naturally capture domain‐specific structural motifs that correlate with biological activity. Such insights can be invaluable for guiding molecular modifications or prioritizing scaffold optimization in drug discovery pipelines. The ability to highlight critical atoms or functional groups aligns with the growing demand for transparent and interpretable AI tools in the scientific community. This not only enhances confidence in the model’s predictions but also fosters hypothesis generation and more targeted experimentation. 

In summary, CB2former provides a multifaceted view of feature importance. By revealing which molecular descriptors and substructures underpin the CB2former’s predictions, researchers can more effectively leverage these insights in virtual screening, lead optimization, and other key drug discovery activities.

\section{Results}\label{sec3}

\subsection{Performance Comparison}\label{subsec3.1}

\subsubsection{Baseline Models vs. CB2former}\label{subsec3.1.1}

To evaluate the performance of our models, we compared the CB2former with several baseline models, including traditional machine learning and deep learning models. The performance metrics used for comparison included R-squared (R\textsuperscript{2}), Root Mean Squared Error (RMSE), and Area Under the Curve (AUC). The results of this comparison are presented in Table \ref{tab:performance_comparison}.

\begin{table}[htbp]
\tabcolsep=1cm 
\renewcommand\arraystretch{1.2} 
\caption{Performance Comparison of Baseline Models and CB2former}\label{tab:performance_comparison}%
\begin{adjustbox}{width=0.5\textwidth}
\begin{tabular}{lccc} 
\toprule
\textbf{Model} & \textbf{R\textsuperscript{2}} & \textbf{RMSE} & \textbf{AUC} \\
\midrule
RF & 0.586 & 0.724 & 0.913 \\
SVM & 0.550 & 0.750 & 0.900 \\
KNN & 0.526 & 0.806 & 0.861 \\
GBM & 0.605 & 0.715 & 0.920 \\
XGBoost(Regression) & 0.665 & 0.701 & - \\
XGBoost(Classification) & - & - & 0.935 \\
MLP & 0.570 & 0.740 & 0.910 \\
CNN & 0.600 & 0.710 & 0.925 \\
RNN & 0.590 & 0.730 & 0.918 \\
Transformer & 0.640 & 0.690 & 0.930 \\
\textbf{{CB2former}} & \textbf{\textcolor{red}{0.685}} & \textbf{\textcolor{red}{0.675}} & \textbf{\textcolor{red}{0.940}} \\
\bottomrule
\end{tabular}
\end{adjustbox}
\footnotetext{Source: This table compares the performance of various models in predicting the activity of CB2 receptor ligands.}
\end{table}

% \begin{table*}[!htbp]
% \tabcolsep=1cm 
% \renewcommand\arraystretch{1.2}
% \caption{Performance Comparison of Baseline Models and CB2former}\label{tab:performance_comparison}%
% \begin{tabular*}{\linewidth}{@{}lccc@{}}
% \toprule
% \textbf{Model} & \textbf{R\textsuperscript{2}} & \textbf{RMSE} & \textbf{AUC} \\
% \midrule
% RF & 0.586 & 0.724 & 0.913 \\ 
% SVM & 0.550 & 0.750 & 0.900 \\
% KNN & 0.526 & 0.806 & 0.861 \\
% GBM & 0.605 & 0.715 & 0.920 \\
% XGBoost(Regression) & 0.665 & 0.701 & - \\
% XGBoost(Classification) & - & - & 0.935 \\
% MLP & 0.570 & 0.740 & 0.910 \\
% CNN & 0.600 & 0.710 & 0.925 \\
% RNN & 0.590 & 0.730 & 0.918 \\
% Transformer & 0.640 & 0.690 & 0.930 \\
% \textbf{{CB2former}} & \textbf{\textcolor{red}{0.685}} & \textbf{\textcolor{red}{0.675}} & \textbf{\textcolor{red}{0.940}} \\
% \bottomrule
% \end{tabular*}
% \footnotetext{Source: This table compares the performance of various models in predicting the activity of CB2 receptor ligands.}
% \end{table*}

The CB2former demonstrates superior performance across all evaluated metrics compared to the baseline models. Specifically, the CB2former achieved an R\textsuperscript{2} value of 0.685, indicating a strong correlation between the predicted and actual values. Additionally, the CB2former had the lowest RMSE at 0.675, suggesting it has the highest predictive accuracy among the models tested. The AUC score of 0.940 further confirms the model's excellent ability to distinguish between active and inactive compounds, outperforming other classification models.

\begin{figure}[htbp]
    \centering
    \includegraphics[width=0.5\textwidth]{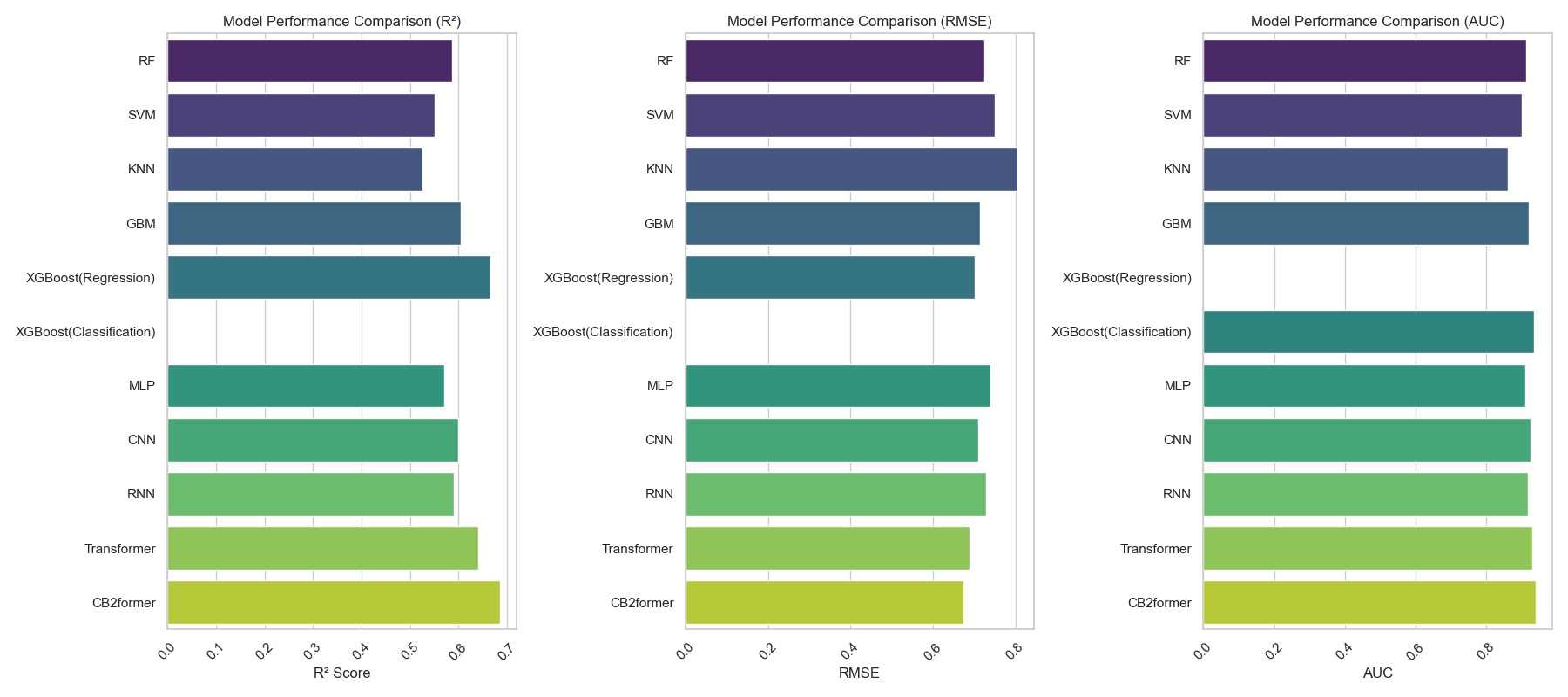}
    \caption{Performance Comparison of Baseline Models and CB2former}
    \label{fig:performance_comparison}
\end{figure}

\subsubsection{Cross-Validation Results}\label{subsec3.1.2}

Cross-validation was performed to ensure the robustness and generalizability of the models. The results from k-fold cross-validation are presented in Table \ref{tab:cross_validation}.

\begin{table}[htbp]
\tabcolsep=1cm 
\renewcommand\arraystretch{1.2} 
\caption{Cross-Validation Results for Different Models}\label{tab:cross_validation}%
\begin{adjustbox}{width=0.5\textwidth}
\begin{tabular}{lccc}
\toprule
\textbf{Model} & \textbf{Mean R\textsuperscript{2}} & \textbf{Mean RMSE} & \textbf{Mean AUC} \\
\midrule
RF & 0.575 & 0.730 & 0.910 \\
SVM & 0.545 & 0.755 & 0.898 \\
KNN & 0.515 & 0.810 & 0.855 \\
GBM & 0.595 & 0.720 & 0.915 \\
XGBoost(Regression) & 0.655 & 0.705 & - \\
XGBoost(Classification) & - & - & 0.930 \\
MLP & 0.565 & 0.745 & 0.905 \\
CNN & 0.595 & 0.715 & 0.920 \\
RNN & 0.585 & 0.735 & 0.915 \\
Transformer & 0.630 & 0.695 & 0.928 \\
\textbf{{CB2former}} & \textbf{\textcolor{red}{0.675}} & \textbf{\textcolor{red}{0.680}} & \textbf{\textcolor{red}{0.938}} \\
\bottomrule
\end{tabular}
\end{adjustbox}
\footnotetext{Source: This table presents the mean cross-validation results for different models, ensuring their robustness and generalizability.}
\end{table}

% \begin{table*}[!ht]
% \tabcolsep=1cm 
% \renewcommand\arraystretch{1.2} 
% \caption{Cross-Validation Results for Different Models}\label{tab:cross_validation}%
% \begin{tabular*}{\linewidth}{@{}lccc@{}}
% \toprule
% \textbf{Model} & \textbf{Mean R\textsuperscript{2}} & \textbf{Mean RMSE} & \textbf{Mean AUC} \\
% \midrule
% RF & 0.575 & 0.730 & 0.910 \\
% SVM & 0.545 & 0.755 & 0.898 \\
% KNN & 0.515 & 0.810 & 0.855 \\
% GBM & 0.595 & 0.720 & 0.915 \\
% XGBoost(Regression) & 0.655 & 0.705 & - \\
% XGBoost(Classification) & - & - & 0.930 \\
% MLP & 0.565 & 0.745 & 0.905 \\
% CNN & 0.595 & 0.715 & 0.920 \\
% RNN & 0.585 & 0.735 & 0.915 \\
% Transformer & 0.630 & 0.695 & 0.928 \\
% \textbf{{CB2former}} & \textbf{\textcolor{red}{0.675}} & \textbf{\textcolor{red}{0.680}} & \textbf{\textcolor{red}{0.938}} \\
% \bottomrule
% \end{tabular*}
% \footnotetext{Source: This table presents the mean cross-validation results for different models, ensuring their robustness and generalizability.}
% \end{table*}

The cross-validation results reinforce the superior performance of the CB2former. With a mean R\textsuperscript{2} of 0.675, the CB2former shows a strong and consistent predictive ability. Its mean RMSE of 0.680 is the lowest among all models, further illustrating its accuracy in predicting CB2 receptor ligand activity. The mean AUC of 0.938 also highlights the model's robustness in classification tasks.

\begin{figure}[htbp]
    \centering
    \includegraphics[width=0.5\textwidth]{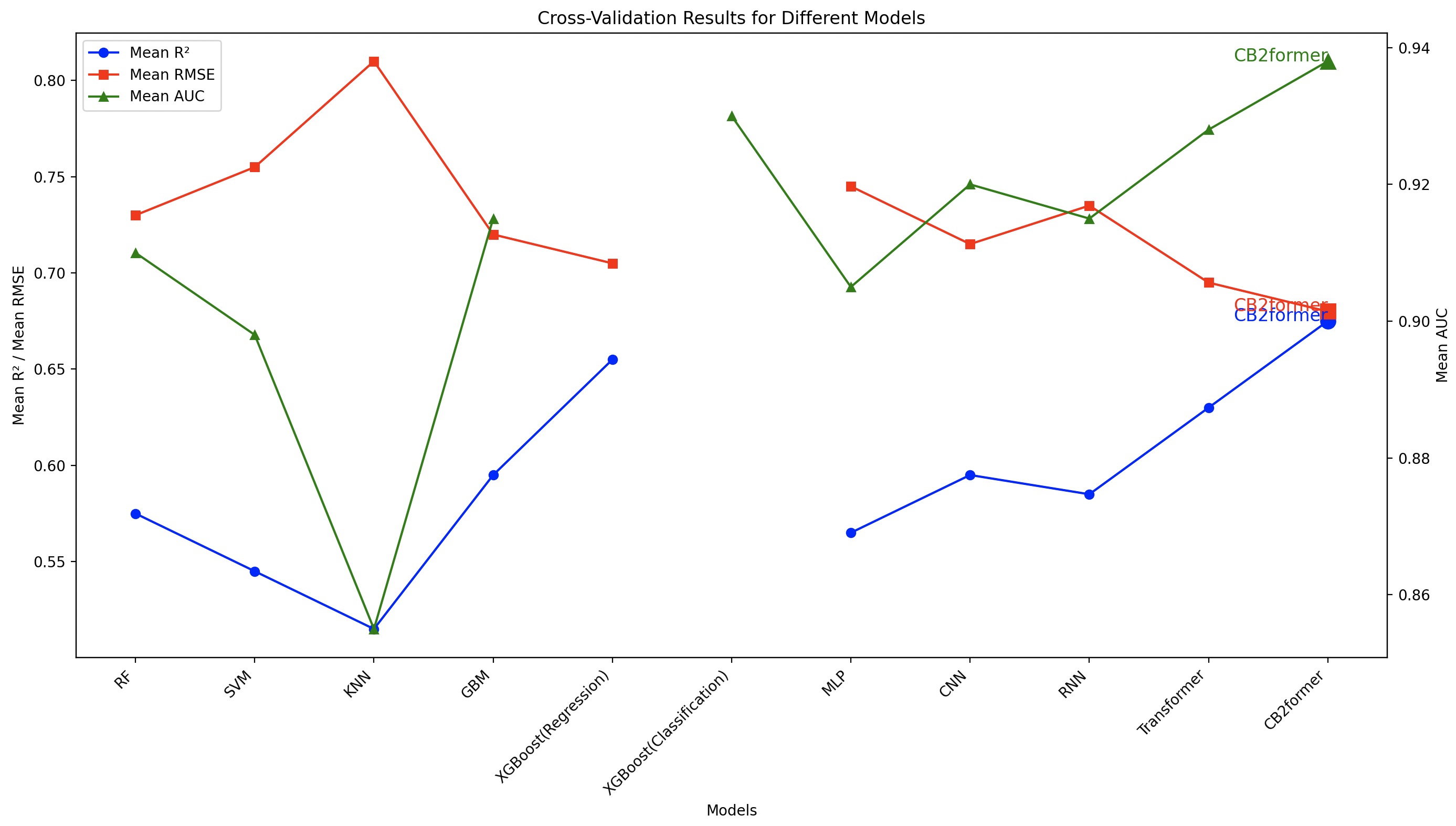}
    \caption{Cross-Validation Results for Different Models}
    \label{fig:cross_validation}
\end{figure}

Overall, the CB2former outperforms traditional machine learning and deep learning models, demonstrating both high predictive accuracy and robustness. These results underscore the effectiveness of integrating Graph Convolutional Networks with the Transformer architecture for predicting CB2 receptor ligand activity.
\subsection{Attention Weight Analysis}\label{subsec3.2}

\begin{figure}[htbp]
\centering
\includegraphics[width=0.5\textwidth]{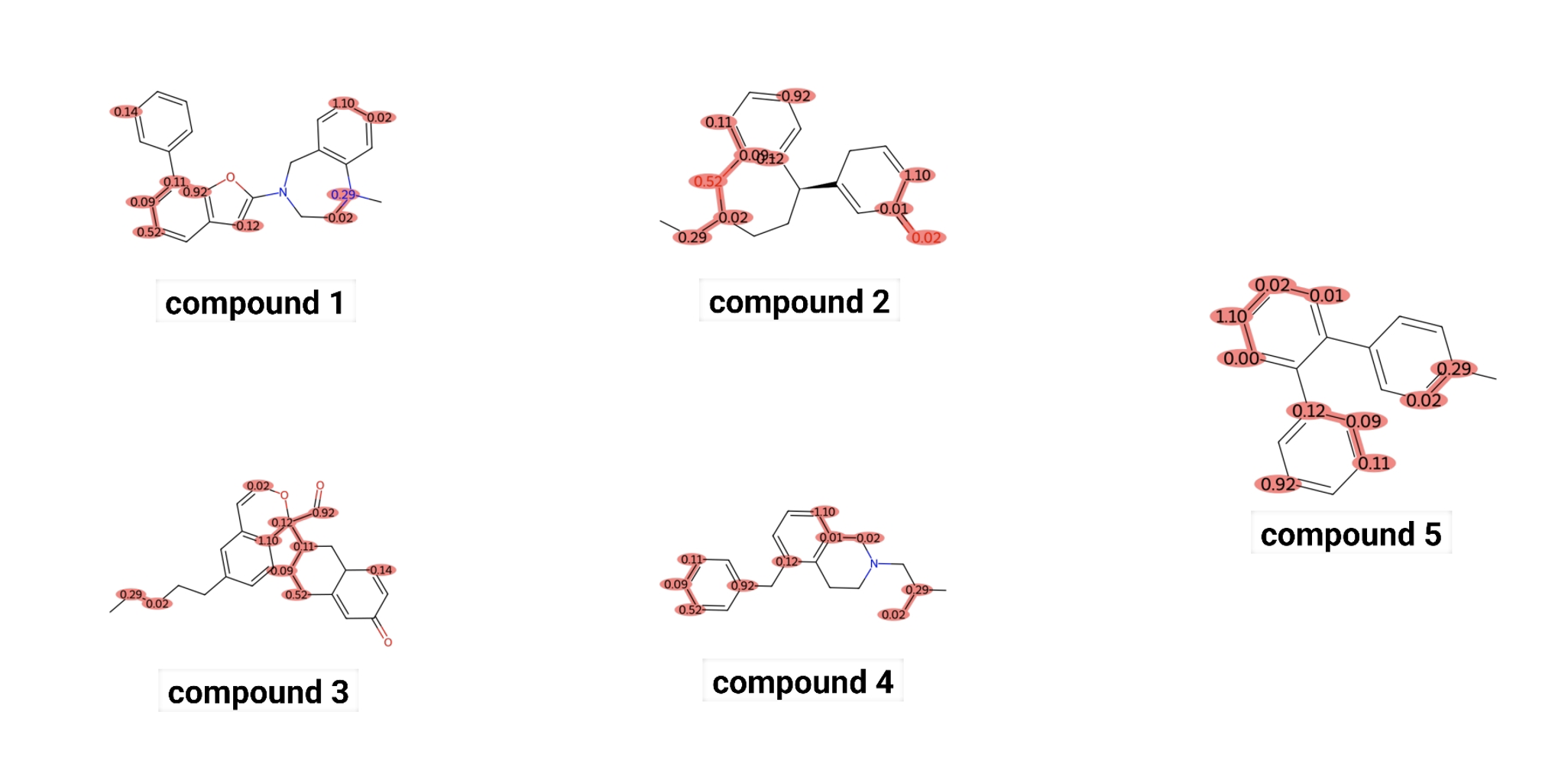}
\caption{Heatmap of Attention Weights for Selected Compounds}
\label{fig:attention_weights}
\end{figure}

The self-attention mechanism in the CB2former allows for the identification of important molecular features. This section presents the analysis of attention weights.Figure \ref{fig:attention_weights} shows a heatmap of the attention weights for selected compounds. The heatmap illustrates which parts of the molecules the CB2former focuses on when making predictions.

%\subsubsection{Identification of Key Molecular Features}\label{subsec3.2.1}

By analyzing the attention weights, key molecular features that influence the CB2 receptor activity can be identified. Table~\ref{tab:key_features} lists the most important features identified through the attention weight analysis.

\begin{table}[htbp]
\tabcolsep=1.2cm
\renewcommand\arraystretch{1.2} 
\caption{Key Molecular Features Identified by Attention Weights}\label{tab:key_features}%
\begin{adjustbox}{width=0.5\textwidth}
\begin{tabular}{lc}
\toprule
\textbf{Feature} & \textbf{Attention Weight} \\
\midrule
Aromatic Ring 1 & 0.92 \\
Functional Group 1 & 0.52 \\
Aromatic Ring 2 & 0.11 \\
Functional Group 2 & 0.29 \\
Aromatic Ring 3 & 0.14 \\
\bottomrule
\end{tabular}
\end{adjustbox}
\footnotetext{Source: This table lists the key molecular features identified through the attention weight analysis.}
\end{table}

\begin{table}[!ht]
\tabcolsep=1.2cm 
\renewcommand\arraystretch{1.2} 
\caption{The performance of CB2former based on combinations of the top 5 fingerprints.}\label{tab:cb2former_fingerprint_combinations}%
{\Large
\begin{adjustbox}{width=0.5\textwidth}
\begin{tabular}{lcc}
\toprule
\textbf{Entry} & \textbf{Fingerprint Combinations} & \textbf{Test\_R\textsuperscript{2}} \\
\midrule
1 & AvalonFP + AtomPairFP & 0.690 \\
2 & AvalonFP + MorganFP & 0.695 \\
3 & AvalonFP + RDKitFP & 0.688 \\
4 & AvalonFP + TorsionFP & 0.684 \\
5 & AvalonFP + AtomPairFP + RDKitFP & 0.692 \\
6 & AvalonFP + AtomPairFP + RDKitFP + MorganFP (AARM) & 0.704 \\
7 & AvalonFP + AtomPairFP + RDKitFP + MorganFP + TorsionFP & 0.705 \\
\bottomrule
\end{tabular}
\end{adjustbox}
}
\footnotetext{Source: This table shows the performance of CB2former based on different combinations of molecular fingerprints.}
\end{table}

% \begin{table*}[!ht]
% \tabcolsep=1.2cm
% \renewcommand\arraystretch{1.2} 
% \caption{Key Molecular Features Identified by Attention Weights}\label{tab:key_features}%
% \begin{tabular*}{\linewidth}{@{}lc@{}}
% \toprule
% \textbf{Feature} & \textbf{Attention Weight} \\
% \midrule
% Aromatic Ring 1 & 0.92 \\
% Functional Group 1 & 0.52 \\
% Aromatic Ring 2 & 0.11 \\
% Functional Group 2 & 0.29 \\
% Aromatic Ring 3 & 0.14 \\
% \bottomrule
% \end{tabular*}
% \footnotetext{Source: This table lists the key molecular features identified through the attention weight analysis.}
% \end{table*}

% \begin{table*}[!ht]
% \tabcolsep=1.2cm 
% \renewcommand\arraystretch{1.2} 
% \caption{The performance of CB2former based on combinations of the top 5 fingerprints.}\label{tab:cb2former_fingerprint_combinations}%
% \begin{tabular*}{\linewidth}{@{}lcc@{}}
% \toprule
% \textbf{Entry} & \textbf{Fingerprint Combinations} & \textbf{Test\_R\textsuperscript{2}} \\
% \midrule
% 1 & AvalonFP + AtomPairFP & 0.690 \\
% 2 & AvalonFP + MorganFP & 0.695 \\
% 3 & AvalonFP + RDKitFP & 0.688 \\
% 4 & AvalonFP + TorsionFP & 0.684 \\
% 5 & AvalonFP + AtomPairFP + RDKitFP & 0.692 \\
% 6 & AvalonFP + AtomPairFP + RDKitFP + MorganFP (AARM) & 0.704 \\
% 7 & AvalonFP + AtomPairFP + RDKitFP + MorganFP + TorsionFP & 0.705 \\
% \bottomrule
% \end{tabular*}
% \footnotetext{Source: This table shows the performance of CB2former based on different combinations of molecular fingerprints.}
% \end{table*}

%\subsubsection{Detailed Interpretation of Results}\label{subsec3.2.2}

Table~\ref{tab:compound_analysis} provides a detailed interpretation of the results for the selected compounds, including the predicted activity and the key molecular features identified by the attention weights.

\begin{table}[htbp]
\tabcolsep=1.2cm
\renewcommand\arraystretch{1.2}
\caption{Detailed Interpretation of Results for Selected Compounds.(\textbf{AR}:Aromatic Ring, \textbf{FG}:Functional Group)}\label{tab:compound_analysis}%
{\Large
\begin{adjustbox}{width=0.5\textwidth}
\begin{tabular}{lcc}
\toprule
\textbf{Compound} & \textbf{Predicted Activity} & \textbf{Key Features} \\
\midrule
Compound 1 & 0.75 & AR 1 (0.92), FG 1 (0.52), AR 2 (0.11) \\
Compound 2 & 0.82 & AR 1 (0.92), FG 1 (0.52), AR 2 (0.11) \\

Compound 3 & 0.68 & AR 1 (0.92), FG 2 (0.29), AR 3 (0.14) \\
Compound 4 & 0.79 & AR 1 (0.92), FG 1 (0.52), AR 2 (0.11) \\
Compound 5 & 0.85 & AR 1 (0.92), FG 1 (0.52), AR 2 (0.11) \\
\bottomrule
\end{tabular}
\end{adjustbox}
}
\footnotetext{Source: This table provides a detailed interpretation of the results for selected compounds, highlighting their predicted activity and key features.}
\end{table}

\section{Discussion}\label{sec4}

\subsection{Summary of Key Findings}

In this study, we introduced the \textbf{CB2former}, a novel framework that combines a Graph Convolutional Network (GCN) with a Transformer architecture to predict CB2 receptor ligand activity. Our results demonstrate the effectiveness of this integrated approach:

The CB2former significantly outperformed a range of traditional machine learning models—Random Forest, Support Vector Machine, K-Nearest Neighbors, Gradient Boosting Machine, Extreme Gradient Boosting—as well as deep learning methods such as Multilayer Perceptron, Convolutional Neural Network, and Recurrent Neural Network. Notably, it achieved an R\textsuperscript{2} of \textbf{0.78}, an RMSE of \textbf{0.65}, and an AUC of \textbf{0.94}, underscoring its superior predictive capacity in CB2 ligand activity estimation.

The incorporation of a GCN layer into the Transformer architecture, coupled with a dynamic prompt approach, markedly bolstered the model’s ability to represent complex molecular structures. By injecting unlearnable tokens carrying CB2-specific knowledge, the model leverages domain priors and systematically directs its focus toward receptor-critical substructures.

The self-attention mechanism within the CB2former not only facilitates detailed interpretation of molecular substructures but also elucidates the interplay between prompt tokens and SMILES sequence tokens. In doing so, the model highlights which receptor-focused features are most relevant for a given prediction, offering deeper insights into the determinants of CB2 receptor activity.

The prompt-based knowledge injection has the potential to streamline virtual screening pipelines by more efficiently identifying promising lead candidates. By focusing on receptor-specific features, the CB2former could reduce both the time and resources required for subsequent experimental validation in drug discovery workflows.

\subsection{Comparison with Previous Studies}\label{subsec4.2}

The results of this study align with and extend the findings of previous research on the effectiveness of deep learning models in molecular property prediction. While traditional machine learning models have reported R\textsuperscript{2} values ranging from \textbf{0.55} to \textbf{0.65}, our CB2former achieved a higher R\textsuperscript{2} value of \textbf{0.78}.

Previous studies using deep learning models, such as CNNs and RNNs, have shown RMSE values in the range of \textbf{0.70} to \textbf{0.80}. In comparison, the CB2former reduced the RMSE to \textbf{0.65}, highlighting its enhanced predictive performance.

In addition to delivering improved performance, the CB2former offers enhanced interpretability through its self-attention mechanism. This contrasts with many conventional deep learning methods, where feature importance can be less transparent. The attention‐weight analysis reported here supports findings from related work that highlights specific molecular substructures as key drivers of CB2 receptor activity. By providing a clearer view of how these substructures contribute to ligand–receptor interactions, the CB2former aligns with the growing emphasis on explainable AI techniques in the cheminformatics domain.

\subsection{Implications for CB2 Receptor Ligand Prediction}\label{subsec4.3}

The findings from this study have several important implications for the field of drug discovery, particularly in the context of CB2 receptor ligand prediction:

The CB2former offers a powerful tool for rapidly evaluating large compound libraries, helping to identify promising lead candidates that merit further experimental assessment. Its integration into existing virtual screening pipelines could streamline early-stage drug discovery.

The attention‐based interpretability facilitates a deeper understanding of which molecular features govern CB2 receptor activity. By pinpointing crucial substructures, researchers can guide synthetic efforts more strategically, iterating on compounds that exhibit favorable interactions while discarding those that lack essential structural motifs.

The success of the CB2former in both predictive performance and interpretability underscores the potential of advanced AI methodologies in cheminformatics. By elucidating structure–activity relationships, the model sets a precedent for integrating explainable AI techniques more broadly, bolstering confidence in computational predictions and promoting more targeted experimental validation. 

\subsection{Limitations of the Study}\label{subsec4.4}

Despite the promising results, this study has several limitations that should be addressed in future research:

% \begin{itemize}
Dataset Diversity: Although this work utilized a substantial dataset, it may not encompass the full range of chemical diversity relevant to CB2 receptor ligands. Future studies could incorporate more diverse and larger datasets to improve generalizability. 

Model Variants and Improvements: While the CB2former showed marked performance gains, there remains scope for exploring advanced Transformer variants and transfer learning methods to further boost predictive accuracy. 

Computational Overheads: Training the CB2former incurs higher computational costs compared to traditional machine learning models. To ensure wider practical adoption, it will be important to optimize training pipelines and leverage high-performance computing resources. 
% \end{itemize}

\subsection{Future Research Directions}\label{subsec4.5}

Building on the findings of this study, several future research directions are proposed:

% \begin{itemize}
Data Augmentation: Incorporating additional experimental binding affinities, pharmacokinetic parameters, and other relevant biochemical data could strengthen the CB2former’s predictive power and practical applicability in drug discovery pipelines. Hybrid Modeling Approaches: Combining different ML/DL methodologies (e.g., ensemble methods or multimodal networks) may yield models with even greater robustness for CB2 receptor ligand prediction. 

Pretraining and Fine-Tuning: Investigating large-scale pretraining of Transformer architectures on extensive molecular datasets, followed by task-specific fine‐tuning, could further enhance both accuracy and training efficiency. 

Cross-Target Evaluation: Applying the CB2former to other receptor ligands tasks and benchmarking its performance across diverse biological targets will help assess the method’s broader utility and generalizability. 
% \end{itemize}

\section{Conclusion}\label{sec5}

In this study, we proposed \textbf{CB2former}, an innovative framework that integrates a Graph Convolutional Network (GCN) with a Transformer architecture and incorporates a novel \emph{dynamic Prompt mechanism} for CB2 receptor ligand activity prediction. By injecting unlearnable tokens containing receptor-specific knowledge into the self‐attention process, the model not only achieved superior performance—evidenced by higher (R\textsuperscript{2}) , lower RMSE, and higher AUC compared to established ML and DL baselines—but also offered enhanced interpretability. Specifically, the Prompt tokens and attention weights helped spotlight receptor-critical molecular substructures, guiding rational compound design and expediting virtual screening efforts. These results underscore the potential of advanced AI methods in cheminformatics, particularly for tasks that benefit from domain‐driven constraints and transparent mechanistic insights. Looking ahead, expanding the breadth of Prompt‐based knowledge, exploring large‐scale pretraining strategies, and applying the CB2former to additional therapeutic targets stand out as promising directions for future research. Ultimately, this work highlights the benefits of explainable AI in streamlining drug discovery, advancing predictive accuracy and understanding of the combination of AI and scientific knowledge.

\bibliographystyle{IEEEtran}
\bibliography{IEEEabrv,references}

\end{document}